\documentclass{article} 
\usepackage{iclr2025_conference,times}

\usepackage{amsmath,amsfonts,bm}

\def\eqref#1{equation~\ref{#1}}

\def\1{\bm{1}}

\DeclareMathAlphabet{\mathsfit}{\encodingdefault}{\sfdefault}{m}{sl}
\SetMathAlphabet{\mathsfit}{bold}{\encodingdefault}{\sfdefault}{bx}{n}

\newcommand{\myparatight}[1]{\smallskip\noindent{\bf {#1}:}}
\usepackage{multirow}
\usepackage{array}
\usepackage[pagebackref=true,breaklinks=true,colorlinks,bookmarks=false,citecolor=blue,linkcolor=blue]{hyperref}
\usepackage{url}
\usepackage{makecell}
\usepackage{graphicx}
\usepackage{subfigure}
\usepackage{booktabs}
\usepackage{amssymb}
\usepackage{bbm}
\usepackage{todonotes}
\usepackage{xcolor}
\usepackage{enumitem}
\usepackage{wrapfig}
\usepackage{xspace}
\usepackage{float}
\usepackage{subcaption}

\hypersetup{pdfborder={0 0 0}}

\newcommand{\func}[1]{{\ttfamily #1}\xspace}

\title{StringLLM: Understanding the String Processing Capability
of Large Language Models}

\author{Xilong Wang$^{1,2}$, Hao Fu$^2$, Jindong Wang$^3$, Neil Zhenqiang Gong$^1$ \\
$^1$Duke University, $^2$Li Auto, $^3$William \& Mary\\
\texttt{\{xilong.wang, neil.gong\}@duke.edu, fuhao8@lixiang.com} \\
\texttt{jwang80@wm.edu}
}

\newcommand\blfootnote[1]{
  \begingroup
\renewcommand\thefootnote{}\footnote{#1}%
  \addtocounter{footnote}{-1}%
  \endgroup
}

\iclrfinalcopy 
\begin{document}

\blfootnote{Work done when Xilong Wang was an intern at Li Auto mentored by Hao Fu.}
\maketitle

\begin{abstract}
String processing, which mainly involves the analysis and manipulation of strings, is a fundamental component of modern computing. Despite the significant advancements of large language models (LLMs) in various natural language processing (NLP) tasks, their capability in string processing remains underexplored and underdeveloped. To bridge this gap, we present a comprehensive study of LLMs' string processing capability. In particular, we first propose \emph{StringLLM}, a method to construct datasets for benchmarking string processing capability of LLMs. We use StringLLM to build a series of datasets, referred to as \emph{StringBench}. It encompasses a wide range of string processing tasks, allowing us to systematically evaluate LLMs' performance in this area. Our evaluations indicate that LLMs struggle with accurately processing strings compared to humans. To uncover the underlying reasons for this limitation, we conduct an in-depth analysis and subsequently propose an effective approach that significantly enhances LLMs' string processing capability via fine-tuning. This work provides a foundation for future research to understand LLMs' string processing capability. Our code and data are available at \url{https://github.com/wxl-lxw/StringLLM}.


\end{abstract}

\section{Introduction}
\label{sec_1}

String processing is one of the most essential tasks in modern computing.
It is mainly involved with string analysis and manipulation such as accessing characters at a specific index (string indexing) or locating a substring within a string (substring searching).
For instance, search engines match user queries to relevant documents using substring searching, with Google processing over 3.5 billion searches per day.
String processing is also crucial for searching through large databases, such as Amazon's over 400 million product listings.
It is also vital for cleaning and normalizing messy text data in machine learning.  Additionally, string processing plays a critical role in  LLM reasoning. For instance,  the chain-of-thought examples in the OpenAI o1 documentation \citep{o1}, such as cipher and crossword problems, frequently involve string processing.
Programming languages like Python offer powerful built-in functions for string manipulation, making it easier for \emph{humans} to solve complex string-related problems.

With the advance of LLMs on natural language processing (NLP) tasks~\citep{o1, gpt4o, achiam2023gpt}, it is intuitive to assume that LLMs should also achieve great performance on string-related problems since strings are texts.
However, the reality is that LLMs often struggle with these seemingly simple challenges.
A popular example \citep{count_straw} shows that GPT-4o \citep{gpt4o} frequently fails in counting the number of ``r"s in the word ``strawberry".
Additionally, the use of LLMs for string processing remains largely unstudied.
Existing works \citep{shin2024large,tan2024can,yehudai2024can,zhou2023instruction} are mostly limited to case studies with only a few string processing tasks, lacking a comprehensive evaluation of LLMs' capabilities and limitations in this domain. Consequently, it remains unclear how well LLMs can handle such tasks, why they fail, and what might improve their performance.

\myparatight{StringLLM} In this paper, we conduct the first comprehensive study toward understanding the string processing capability of LLMs. One core challenge is how to create large representative datasets with a diverse set of string processing tasks, different types of strings, and guaranteed ground truth answers. To address the challenge, we propose \emph{StringLLM}, the \emph{first} method to construct datasets for benchmarking string processing capability of LLMs. 
StringLLM begins by manually collecting a set of fundamental string processing tasks called \emph{atomic tasks}. Then, StringLLM combines these atomic tasks to create more intricate ones, referred to as \emph{composite tasks}. Given a task, StringLLM generates question templates for it, which are then used to derive question inputs. 
Finally, StringLLM obtains the ground truth answer for a question input by executing the Python code of the corresponding task, generating question-answer pairs for string processing. 


\myparatight{Crafting new datasets} We use StringLLM to construct a series of datasets, referred to as \emph{StringBench}. 
These datasets include 1,511 string processing tasks spanning a broad range of applications, and three common types of strings: hash strings, multilingual natural language strings, and random strings. StringBench allows us to thoroughly assess the ability of LLMs to process strings across various scenarios. 

\myparatight{Benchmarking string processing capabilities of LLMs} Building on StringBench, we present the \emph{first} systematic evaluation of LLMs on string processing tasks, using three prompting strategies: raw instructions, Chain of Thought (CoT) \citep{wei2022chain}, and Program of Thought (PoT) \citep{chen2022program}. Our comprehensive experimental results demonstrate that: 1) LLMs struggle with string processing compared to human capability. In particular, they achieve a maximum of 48.89\% accuracy using raw instructions; 2) LLMs' performance varies across datasets, revealing significant disparities in their ability to process different types of strings. Specifically, random strings are the most challenging, with accuracy peaking at 43.94\% using raw instructions; and 3) Prompt engineering significantly improves performance. Some LLMs show over a 20\% improvement when using PoT compared to raw instructions, offering valuable insights for designing better solutions.

\myparatight{Understanding why LLMs struggle with string processing} We conduct the \emph{first} in-depth analysis to investigate why LLMs struggle with string processing. Our analysis reveals that: 1) Tokenization fails to split strings into individual characters, resulting in a lack of character-level understanding in LLMs; and 2) Token embedding lacks character-level information such as token length information, further highlighting LLMs' limited character-level comprehension of strings.
Our analysis and \citet{yehudai2024can} show that Transformers have limited ability in solving string processing tasks. To address this, we propose an effective solution to enhance LLMs' performance in string processing, without altering the architecture of Transformers. Utilizing our well-constructed StringBench, we conduct supervised fine-tuning on three different open-source LLMs. Our fine-tuned models improve average test accuracy of our datasets by at least \textbf{38.80\%}, compared to the best-performing prompt engineering technique, PoT. We then evaluate the foundational capabilities of our fine-tuned models on three general-purpose benchmarks. The results show that the string processing capabilities of our fine-tuned models are significantly enhanced without substantially degrading their foundational capabilities. Specifically, the three fine-tuned LLMs sacrifice at most 1.35\% on average across the three general-purpose benchmarks.

    

\section{Related Work}
\myparatight{LLMs' string processing capability is underexplored} LLMs have advanced significantly in recent years, demonstrating impressive capabilities across diverse NLP tasks, such as reasoning \citep{wei2022chain, chen2022program}, coding \citep{roziere2023code, zhu2024deepseek}, and instruction following \citep{ouyang2022training}. However, their string processing capabilities remain understudied. \citet{shin2024large} explored LLMs' ability to handle character-level tasks as opposed to token-level tasks. Their study revealed that LLMs struggle with simple character-level operations (e.g., character insertion, deletion, reordering), which humans can perform with ease. In contrast, LLMs' performance on token-level tasks (e.g., sentence retrieval, word insertion) is relatively stronger, highlighting a discrepancy between token and character-level understanding. \citet{zhou2023instruction} proposed a benchmark to evaluate the instruction-following capabilities of LLMs, including some string processing tasks. However, the tasks are few and lack comprehensiveness, limiting their use to case studies. \citet{tan2024can} further noted that LLMs often struggle with basic counting tasks, such as generating a paragraph with a specific word count and then correctly identifying that number. Correspondingly, \citet{yehudai2024can} discussed strategies to address this counting problem in Transformers. However, these works are limited to case studies focused on single or a few string processing tasks, lacking a comprehensive evaluation. Furthermore, they do not provide an in-depth analysis of why LLMs struggle with accurately processing strings, nor do they offer solid experimental evidence to support their conclusions. Consequently, they fail to propose effective solutions to enhance string processing capability of LLMs. 

\begin{figure}[t]
\begin{center}
\includegraphics[width=0.9\linewidth]{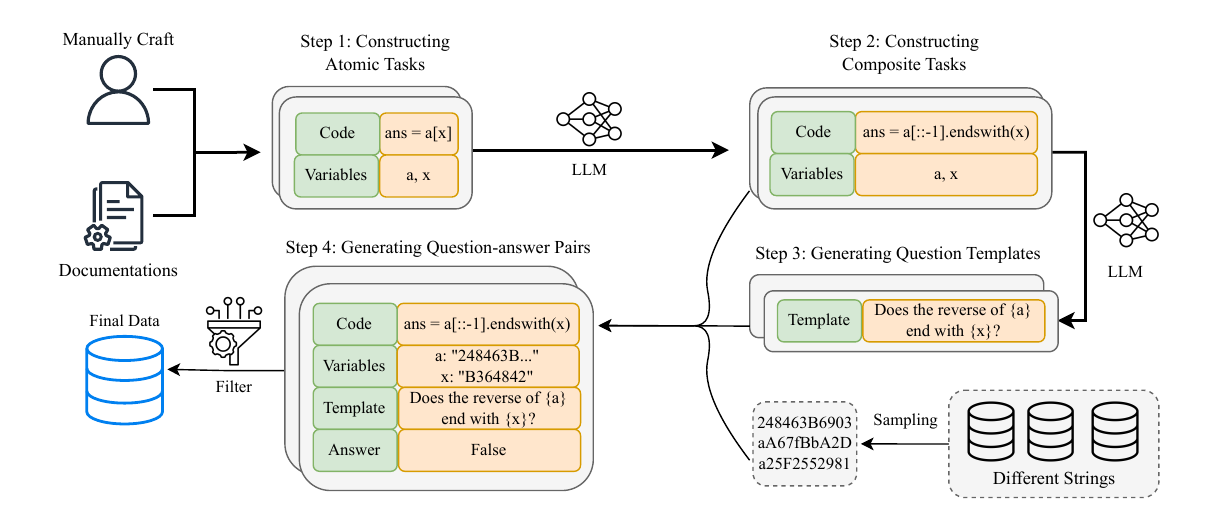}
\end{center}
\caption{Overview of how StringLLM builds the benchmark datasets. \emph{Code} represents Python code expression for each string processing task, \emph{Template} denotes the generated question template for each task, \emph{Variables} are the placeholders within both \emph{Code} and \emph{Template}, and \emph{Answer} refers to the groundtruth for each sample.
}
\label{fig:stringllm}
\end{figure}

\section{Constructing Datasets via StringLLM}
\label{sec:benchmark}



\subsection{Overview}
An overview of StringLLM is shown in Figure \ref{fig:stringllm}. We begin by creating tasks. Specifically, we first create tasks that are fundamental and cannot be broken down further, referred to as \emph{atomic tasks}. These are either manually designed by us or sourced from public programming language documentation. Once we have these atomic tasks, we combine multiple of them into more complex tasks, referred to as \emph{composite tasks}. Using both atomic and composite tasks, we generate natural language question-answer pairs for LLMs, resulting in a series of datasets called \emph{StringBench}. It includes three distinct datasets, each designed to query LLMs in different ways: 
\begin{itemize}[leftmargin=1em]
\setlength\itemsep{0em}
    \item \textbf{Multilingual}: Queries LLMs to process multilingual natural language strings.
    \item \textbf{Hash}: Queries LLMs to process strings encoded by different hash functions.
    \item \textbf{Random Strings}: Queries LLMs to process random strings composed of printable characters.
\end{itemize}
The three datasets cover diverse usage of LLMs, including natural language tasks and code-related tasks. The Multilingual dataset focuses on natural language tasks with special attention to multiple languages. Hash strings represent a typical type of strings in code-related tasks. To cover a wider range of characters, we also consider random strings of all printable characters.
Manually constructing datasets of this nature is extremely time-consuming and impractical. We leverage GPT-4o \citep{gpt4o} to assist in the data construction process.
We avoid direct use of GPT-4o for string processing, as we will show in Section \ref{sec: exp} that it struggles with string-related tasks. Instead, we make use of its coding abilities, including code generation, understanding, and summarization. We take code as a proxy to produce accurate datasets. Previous study \citep{gpt4o} has shown that GPT-4o performs exceptionally well in these areas, achieving a score of 90.2\% on the HumanEval benchmark \citep{chen2021evaluating}. This highlights its remarkable capabilities in code-related applications, enabling us to use it for data construction.

\subsection{Constructing String Processing Tasks}

\myparatight{Atomic string processing tasks} We begin by thoroughly examining the built-in function documentation \footnote{\url{https://docs.python.org/3/library/stdtypes.html\#string-methods}} of Python to identify atomic string processing tasks.
Specifically, we focus on distinguishing unique functions while filtering out those that are redundant or exhibit overlapping functionalities. For example, the \func{splitlines()} function, which is designed for splitting strings at line boundaries, is obviously a specialized case of the more general \func{split()} function. Therefore, we integrate \func{splitlines()} into the broader \func{split()} task and exclude it from separate consideration. Following this, a set of atomic string processing tasks is obtained. However, this initial set proves to be insufficient for our comprehensive needs. Many common tasks are not encapsulated within the Python built-in functions. To address this gap, we manually develop an additional set of atomic tasks. These include fundamental tasks such as indexing, slicing, and concatenation. They are frequently used but are not explicitly provided as built-in functions. Once we have determined the complete set of atomic tasks, we proceed to formally express each task in Python code. We manually create Python scripts that perform these tasks using placeholder variables. These scripts then serve as templates for constructing composite tasks (see Table \ref{table:atomic_tasks} in Appendix for examples).

By manually reviewing the built-in function documentation and crafting an additional set of atomic tasks, we ensure that these atomic tasks are representative and can compose many diverse composite tasks. Moreover, by filtering out tasks that are redundant or exhibit overlapping functionalities, we avoid the selected atomic tasks having conflicted meanings that could form bad composite tasks.

\myparatight{Composite string processing tasks} Building on atomic tasks, we iteratively generate more complex string processing tasks. In each iteration, we randomly shuffle and then input all atomic tasks into GPT-4o, combining them into composite tasks. Specifically, we prompt GPT-4o to generate Python code for composite tasks based on the code of atomic tasks. This approach fully leverages GPT-4o’s Program of Thought (PoT) capability, instead of using it for direct string processing. As a result, we generate a total of 1,462 composite tasks, covering a wide range of complicated string processing scenarios commonly used in practice. 

\subsection{Generating Question-answer Pairs for Each Task}

\myparatight{Constructing question templates}
After obtaining the atomic and composite tasks, we once again use GPT-4o to generate question templates for these tasks. As illustrated in Figure \ref{fig:stringllm}, a question template describes the task with placeholders. Given the Python code, GPT-4o is prompted to create a template that asks how the task described by the code would be accomplished. This ensures that the question templates are aligned with the tasks defined by the Python code. 
\begin{wraptable}{r}{9.5cm}
    \centering
    \caption{Dataset statistics.}
    \label{tab:data_detail}
    \resizebox{.65\textwidth}{!}{
    \begin{tabular}{cccc}
    \toprule
        Dataset & Atomic Tasks  & Composite Tasks & Question-answer Pairs\\
    \midrule
    
     Multilingual &  49 & 1,462 & 22,161\\
     
      Hash  &  49 & 1,462 & 18,441\\
      
      Random String &  49 & 1,462 & 21,159 \\
    \bottomrule
    \end{tabular}
    }
    
\end{wraptable}
Thus, such templates can effectively guide LLMs to perform the required tasks. This procedure leverages GPT-4o's capability of code summarization, without requiring it to process strings directly. As with previous steps, this approach skillfully exploits GPT-4's strengths while minimizing its limitations. To promote diversity, we generate three paraphrased question templates for each task. Specifically, to ensure the quality of the templates, we first instruct GPT-4o to generate 10 templates. Then, we have GPT-4o itself rank these generated templates and select the top three.

\myparatight{Generating question-answer pairs based on the template} 
Once the question templates are generated, we replace the placeholders in the templates with specific strings to be processed. This procedure transforms the templates into question inputs. To promote the diversity of our dataset, we create three types of strings to each question template. This results in StringBench, containing three distinct datasets: 
\begin{itemize}[leftmargin=1em]
\setlength\itemsep{0em}
    \item \textbf{Multilingual:} We randomly sample strings from the Flores-200 dataset \citep{costa2022no}, which is a machine translation dataset including data in 200 languages. 
    \item \textbf{Hash:} We first randomly sample strings from the Flores-200 dataset. For each string, we randomly select a hash function from a set of 10 different functions to encode it. The 10 used cryptographic hash functions are listed in Table \ref{table:hash_lengths} in Appendix.
    \item \textbf{Random String:} We randomly sample strings using all printable characters in Python, including letters, numbers, punctuation, escape characters, and other special symbols.
\end{itemize}

Finally, we execute the Python code to generate the ground truth answers for all datasets. This ensures a reliable and accurate reference for both training and evaluation purposes. 
The guaranteed ground truth data is a significant strength of our approach. Unlike other datasets that may suffer from inconsistencies or ambiguities in labeling, our ground truth provides a definitive reference, improving the reliability of evaluations. This rigorous standard allows us to measure LLMs' performance with high precision and offers a robust foundation for training LLMs to achieve better performance. By leveraging this guaranteed ground truth, we ensure that our datasets serve as a valuable benchmark for both research and practical applications.
Statistics of our datasets are shown in Table \ref{tab:data_detail}, and randomly selected question-answer pairs from each dataset are presented in Table \ref{tab:example}. 

\begin{table}[h]
    \centering
    \caption{Randomly selected question-answer pairs from our datasets. 
    }
    \label{tab:example}
    \resizebox{\textwidth}{!}{
    \begin{tabular}{@{}l|l|c@{}}
    \toprule
       Dataset  & Question & Answer\\
    \midrule
    \toprule

        Multilingual & \begin{tabular}[c]{@{}l@{}}Retrieve the first 29 characters from string \func{"Felicia, rakamboita dutu}\\\func{rechikamu chechina pa Sikero yeSaffir-Simpson Hurricane,}\\\func{yakaneta kusvika kuderera kwetropical ichizopra neChipiri".}\end{tabular} & \begin{tabular}[c]{@{}l@{}}\func{"Felicia,} \\\func{rakamboita}\\\func{dutu rech"}\end{tabular}\\
    \midrule
        Hash & \begin{tabular}[c]{@{}l@{}}Get the index of the first instance of \func{"dc"} in string \func{"c11c8a595476dcde4f91a8}\\ \func{dce2acaba2"}.\\\end{tabular} & 12\\
    \midrule
            Random String & \begin{tabular}[c]{@{}l@{}}Take every 4th character from the start of string \func{"N/5qe!wj8U*8dvsN/am'UGfN}\\ \func{/A=n+\%\$5)3HA?d\#Jn\&F4\&,(WG-p:1Vw]"} up to the 50th character.\end{tabular} & \begin{tabular}[c]{@{}l@{}}\func{"Ne8d/U/}\\\func{+)?n\&G"}\end{tabular}\\

    \bottomrule
    \end{tabular}
    }
\end{table}

\subsection{Post-processing}
We filter out duplicate and low-quality data after the above steps. Duplicate code and question templates are removed in Steps 2 and 3 in Figure \ref{fig:stringllm}. All Python code is validated through execution, and any code that results in execution errors is also removed.

\section{Benchmarking LLMs' String Processing Capabilities}
\label{sec: exp}
\subsection{Experimental Setup}
\label{exp_setup}
\myparatight{LLMs} We use  the following LLMs for our evaluation: GPT-4-Turbo \citep{gpt4turbo}, GPT-4o \citep{gpt4o}, GPT-3.5 \citep{gpt3.5}, DeepSeek-Chat \citep{deepseekai2024deepseekv2strongeconomicalefficient}, Gemma-2-9b \citep{gemma2}, Llama-3.1-8B \citep{llama-3.1-8b}, Mistral-7B-v0.3 \citep{mistral-7b}, Mathstral-7B-v0.1 \citep{mathstral-7b}, Codestral-22B-v0.1 \citep{codestral} and DeepSeek-Coder \citep{zhu2024deepseek}. Table \ref{tab:2} in Appendix shows more details of the models.

\myparatight{Training-test split of our datasets}
For the test sets, we randomly split 20\% of the data from each of the three datasets—Multilingual, Hash, and Random String. We ensure that the test sets cover the full range of 1,511 string processing tasks in our datasets. This approach guarantees that the diversity and complexity of these tasks are well-represented in the test sets, allowing a comprehensive evaluation of the model's performance. The remaining 80\% of our datasets is used as the training sets for our experiments on fine-tuning LLMs in Section~\ref{sec:fine-tuning}.

\myparatight{Prompt engineering} In our experiment, we apply three prompt engineering techniques to evaluate the performance of LLMs. 1) \textbf{Raw instructions:} The questions are input directly to the LLMs without any additional guidance, allowing LLMs to choose their own methods for processing the strings.
2) \textbf{Chain of Thought (CoT)} \citep{wei2022chain}: Prompts LLMs to generate step-by-step solutions for string processing tasks. By breaking down problems into smaller steps, this technique helps LLMs handle complex reasoning tasks more systematically.
3) \textbf{Program of Thought (PoT)} \citep{chen2022program}: Prompts LLMs to generate responses mainly in programming language. This technique structures the reasoning process of LLMs in a programmatic manner. Thus, it enables LLMs to perform intricate string processing tasks with greater precision.


\myparatight{Evaluation metrics}
To evaluate the performance of LLMs on string processing tasks, we use  \emph{Accuracy (Acc)}. It measures the correctness of LLMs' responses to input questions. As mentioned in Section \ref{sec:benchmark}, we execute the Python code of each task to obtain the ground truth answer. Hence, due to the deterministic nature of these answers, correctness is determined by exact matches between the model's final output and the ground truth answer.

\myparatight{Human study} We conducted a human study to evaluate the string processing capabilities of LLMs in comparison with humans. We ask human annotators to perform all string processing tasks in the test set, either manually or by writing Python code. 


\begin{table}[t]
    \centering
    \caption{Accuracy for string processing capability of humans and different LLMs.}
    \label{tab:eval_all}
    \resizebox{.75\textwidth}{!}{
    \begin{tabular}{@{}l c  c c c c c}
\toprule
LLM & Method  & Multilingual & Hash & Random String & AVG \\
\midrule
       \multirow{3}{*}{GPT-4o} & Raw Inst.  & 43.09 &  48.89 & 43.94 & 45.31  \\
                  & CoT & 50.05 &  52.01 &  45.63  & 49.23 \\
                  & PoT & 69.19 & 68.00  &  49.06  & 62.08 \\
       \midrule
                  \multirow{3}{*}{GPT-4-Turbo} & Raw Inst. & 35.49 & 41.61  & 33.87 & 36.99  \\
                  & CoT & 43.98 & 48.10  &  43.09  & 45.06 \\
                  & PoT & 66.50 & 71.01  &  48.74  & 62.08 \\
       \midrule
      \multirow{3}{*}{GPT-3.5}& Raw Inst.  & 8.36 & 15.93  & 13.60 & 12.63  \\
                  & CoT & 30.81 & 29.94  &  24.77  & 28.51 \\
                  & PoT & 42.63 & 42.62  &  25.26  & 36.84 \\
        \midrule
        \multirow{3}{*}{DeepSeek-Coder} & Raw Inst.  & 17.79 &  21.34 & 13.97 & 17.70  \\
                  & CoT & 22.87 & 29.72  &  22.87  & 25.15 \\
                  & PoT & 54.85 &  57.46 &  29.42  & 47.24 \\
       \midrule
\multirow{3}{*}{DeepSeek-Chat}& Raw Inst.  & 4.26 & 3.50  & 1.90 & 3.22  \\
                  & CoT & 7.89 & 9.99  &  7.30  & 8.39 \\
                  & PoT & 7.12 & 7.76  &  2.39  & 5.76 \\
       \midrule
\multirow{3}{*}{Gemma-2-9b}& Raw Inst.  & 21.49 &  21.01 & 14.72 & 19.07  \\
                  & CoT & 23.81 &  24.67 &  18.84  & 22.44 \\
                  & PoT & 55.34 & 52.71  &  14.49  & 40.85 \\
       \midrule
\multirow{3}{*}{Mistral-7B-v0.3}& Raw Inst.  & 4.54 & 3.83  & 2.67 & 3.68  \\
                  & CoT & 10.56 &  8.77 &  6.02  & 8.45 \\
                  & PoT & 32.54 & 31.57  &  14.07  & 26.06 \\
       \midrule
\multirow{3}{*}{Mathstral-7B-v0.1}& Raw Inst.  & 4.85 &  5.81 & 2.08 & 4.25  \\
                  & CoT & 13.25 & 15.72  &  10.07  & 13.01 \\
                  & PoT & 21.40 & 25.95  &  12.31  & 19.89\\
       \midrule
\multirow{3}{*}{Codestral-22B-v0.1}& Raw Inst.  & 19.03 &  15.99 & 12.52 & 15.85  \\
                  & CoT & 16.08 & 24.46  &  10.56  & 17.03 \\
                  & PoT & 23.92 & 14.39  &  15.75  & 18.02 \\
       \midrule
\multirow{3}{*}{Llama-3.1-8B}& Raw Inst.  & 11.87 & 16.61  & 10.30 & 12.93  \\
                  & CoT & 19.85 & 22.04  &  17.13  & 19.67 \\
                  & PoT & 39.35 & 42.62  &  21.79  & 34.59 \\
\bottomrule
\toprule
\multirow{2}{*}{Human}& Manual & 42.27 & 95.74 & 93.55 & 77.19 \\
                        & Python  & 98.12 & 97.99 & 98.61 & 98.24\\
\bottomrule
    \end{tabular}
    }
    
\end{table}

\subsection{Experimental Results}

\myparatight{LLMs struggle to process strings compared to humans} The experimental results in Table \ref{tab:eval_all} reveal that all LLMs struggle with string processing tasks across all datasets and prompt engineering techniques. For example, when using raw instruction prompts, even the best-performing LLMs, GPT-4o and GPT-4-Turbo, achieve only 36.99\% and 45.31\% Acc on average, respectively. Other LLMs perform significantly worse. For instance, GPT-3.5 reaches an average Acc of just 12.63\%, which is 32.68\% lower than its updated version GPT-4o. LLMs with much fewer parameters exhibit even more significant performance drops, with average Acc below 20\%. Specifically, Mistral-7B-v0.3 and DeepSeek-Chat exhibit particularly poor performance. They achieve average Acc of only 3.68\% and 3.22\%, respectively. In contrast, humans demonstrate near-perfect performance across all datasets. They do struggle to process strings manually, as they cannot identify and segment Multilingual characters. For instance, in the case of the Arabic language, non-native speakers may find it challenging to process such strings due to the difficulty of identifying individual characters. However, when using Python code to process strings, humans achieve an average Acc of 98.24\%. This highlights the limitations of LLMs compared to human capability.

\myparatight{Random strings are harder to process} As shown in Table \ref{tab:eval_all}, LLMs exhibit significant performance variation across our datasets. In general, LLMs perform better on the Hash dataset compared to the Random String and Multilingual datasets. For instance, with raw instruction prompts, GPT-4o achieves 48.89\% Acc on the Hash dataset. This is 5.8\% higher than its performance on the Multilingual dataset and 4.95\% higher than on the Random String dataset. Overall, 6 out of 10 LLMs perform best on the Hash dataset, indicating that LLMs excel in processing hash strings. Additionally, 8 out of 10 LLMs show their worst performance on the Random String dataset when using raw instruction prompts. This finding suggests that LLMs lack expertise in processing random strings. One possible reason is that LLMs are not trained on data containing random strings, making it an out-of-distribution problem.

\myparatight{Prompt engineering improves performance} Table \ref{tab:eval_all} shows that LLMs perform largely better with PoT and CoT than with raw instructions, where all 10 LLMs exhibit their lowest average Acc. In contrast, when prompted with PoT, LLMs consistently deliver the best results. Specifically, 9 out of 10 LLMs show their highest performance. For example, the DeepSeek-Coder achieves an average Acc of 47.24\% across the three datasets when using PoT. It is 22.09\% and 29.54\% higher than with CoT and raw instructions, respectively. This is likely because, compared to CoT, PoT structures the reasoning process more hierarchically and incorporates programming to solve problems. As a result, it is better suited for precise character-level manipulation required in string processing. Examples of GPT-4o solving a string processing task using different prompt engineering techniques are shown in Figure \ref{fig: case_study_1} - \ref{fig: case_study_2} in Appendix. These examples further highlight PoT's advantage over CoT and raw instructions in string processing.

\section{Understanding Why LLMs Struggle with String Processing}
\label{sec:analysis}

\subsection{Tokenization Cannot Split Strings into Characters}
Tokenization is the process of breaking down texts into smaller units called tokens, which can be words, subwords, or characters.
This process is fundamental to how LLMs process and understand text.
However, LLMs lose character-level details of input, when it is tokenized.
When an input text is tokenized into subwords or words, LLMs may lose the granular character-level details of the input.
This can lead to issues for string processing tasks, which require precise character-level manipulation or analysis. For instance, checking if a string is a palindrome (reads the same backward as forward) can be problematic due to tokenization. Taking a simple example, the phrase \func{"A man a plan a canal Panama"} should be recognized as a palindrome when ignoring spaces and case differences. However, the tokenizer might split it into \func{["A", "man", "a", "plan", "a", "canal", "Panama"]}. As a result, LLMs may fail to perform the check correctly, since they may treat each word as a separate token instead of considering the entire character sequence.

Table \ref{tab:token_ratio} shows ratios of actual length to the number of tokens for sampled strings in our datasets, when tokenized by different LLMs. From Table \ref{tab:token_ratio}, we can draw the following conclusion: 
\begin{itemize}[leftmargin=1em]
\setlength\itemsep{0em}
    \item  For the Multilingual dataset, as shown in Table \ref{tab:token_ratio}, all three LLMs show the highest ratio. This is because in multilingual natural language strings, words and subwords remain indivisible during tokenization. Consequently, tokenization does not break down the strings into finer fragments. This could contribute to lower Acc in such scenarios.
    \item For the Hash dataset, tokenization breaks the string into finer fragments. As shown in Table \ref{tab:token_ratio}, two out of three LLMs display the lowest ratio. Furthermore, hash strings typically consist of alphanumeric characters without spaces or other natural language patterns. They also have a fixed length, making them relatively structured. Additionally, they generally do not contain punctuation or escape characters, further simplifying their structure. In this context, hash strings are easier for LLMs to process and analyze, leading to higher Acc. 
    \item 
    For the Random String dataset, tokenization segments the strings into smaller units as well. However, these strings often contain escape characters, punctuation, and other special characters. These elements add complexity to the structure of random strings, making it challenging for LLMs to analyze and understand them accurately. 
\end{itemize}

\begin{wraptable}{r}{7.4cm}
    \centering
    \caption{Ratios of actual length to the number of tokens for sampled strings in our datasets, when tokenized by different LLMs. 
    }
    \resizebox{.53\textwidth}{!}{
    \begin{tabular}{@{}l c  c c c}
\toprule
LLM & Multilingual & Hash & Random String \\
\midrule

                  Gemma-2-9b  & 2.53 & 1.12  &  1.30    \\
       
       Mistral-7B-v0.3  & 1.73 & 1.06 & 1.15   \\
       
      Llama-3.1-8B & 2.13 & 1.67 & 1.32  \\
\bottomrule
    \end{tabular}
    }
    \label{tab:token_ratio}
\end{wraptable}
Correspondingly, as shown in Table \ref{tab:eval_all}, LLMs perform best on the Hash dataset due to the simpler, more structured nature of the strings. For the Multilingual dataset, performance is lower because tokenization preserves the integrity of words and subwords. Consequently, it could be more difficult for LLMs to establish character-level understanding of such strings. For the Random String dataset, the added complexity from special characters further challenges LLMs, resulting in the lowest Acc.

\subsection{Tokenization Makes LLMs Lack Character-Level Understanding}
As stated in the previous section, tokenization typically splits strings into word-level or subword-level tokens instead of individual characters. However, existing studies do not provide sufficient evaluation to confirm that these tokens lack character-level information. In this section, we aim to further demonstrate that current tokenization, which cannot perform such fine-grained splitting, makes LLMs indeed lack character-level information.
Table \ref{tab:eval_spaces} shows the string processing capability of different LLMs, where white spaces are inserted between all characters of strings. This insertion of white spaces aims to segment every character in strings, avoiding the use of word-level or subword-level tokens, thereby introducing potential character-level information to LLMs.

\begin{table}[htbp]
    \centering
    \caption{Accuracy for string processing capability of different LLMs, where white spaces are inserted between all characters of strings.}
    \label{tab:eval_spaces}
    \resizebox{.75\textwidth}{!}{
    \begin{tabular}{@{}l c  c c c c c}
\toprule
LLM & Method  & Multilingual & Hash & Random String & AVG \\
\midrule
\multirow{3}{*}{Gemma-2-9b}& Raw Inst.  & 20.48 &  21.23 & 14.31 &  18.67 \\
                  & CoT & 24.22 & 25.16 & 20.07 & 23.15
 \\
                  & PoT & 55.69 & 52.73 & 18.98 & 42.47
 \\
       \midrule
\multirow{3}{*}{Mistral-7B-v0.3}& Raw Inst.  & 4.97 & 3.99 & 2.79 & 3.92
 \\
                  & CoT & 12.45 & 10.53 & 7.34 & 10.11

\\
                  & PoT & 32.28 & 32.67  &  15.32  & 26.76 \\
       \midrule

\multirow{3}{*}{Llama-3.1-8B}& Raw Inst.  & 12.83 & 16.56  & 9.76 &  13.05 \\
                  & CoT & 20.64 & 23.21  &  19.20  & 21.02 \\
                  & PoT & 42.01 & 43.30  &  24.74  & 36.68 \\
\bottomrule

    \end{tabular}
    }
    
\end{table}

Compared to Table \ref{tab:eval_all}, the performance of LLMs is improved, indicating that the word-level or subword-level tokens make LLMs lack character-level understanding. In contrast, augmenting the tokenizer with finer-grained character-level segmentation could be a promising direction for improving LLM performance. 
\subsection{LLMs' Understanding of Strings is Limited}
Our experimental results have demonstrated that tokenization cannot split strings into individual characters, and word-level or subword-level tokens do not include sufficient character-level information. Considering the architecture of Transformers, tokenization is the starting point, and its limitations can lead to further errors. Consequently, the attention mechanism and entire Transformer architecture of LLMs cannot effectively analyze character-level information, leading to a lack of fundamental understanding of strings.

\section{Improving String Processing Capability via Fine-tuning}
\label{sec:fine-tuning}


\begin{table}[t]
    \centering
    \caption{Acc for string processing capability of different LLMs before and after fine-tuning, when prompted with PoT. 
    }
    \resizebox{.8\textwidth}{!}{
    \begin{tabular}{@{}l l  c c c c c}
\toprule
LLM &   & Multilingual & Hash & Random String & Avg. Change \\
\midrule

                  \multirow{2}{*}{Gemma-2-9b} & Before  & 55.34 &  52.71 & 14.49    & \multirow{2}{*}{+ 38.80} \\
                  & After  & 82.49 & 87.98  &  68.46 & \\
       \midrule
       \multirow{2}{*}{Mistral-7B-v0.3} & Before & 32.54 & 31.57 & 14.07 & \multirow{2}{*}{+ 56.51}  \\
       & After & 83.24 & 90.28 &  74.18 &  \\
       \midrule
      \multirow{2}{*}{Llama-3.1-8B}& Before & 39.35 & 42.62 & 21.79 & \multirow{2}{*}{+ 42.87} \\
 & After & 79.83 & 85.64 & 66.91 & \\
\bottomrule
    \end{tabular}
    }
    \label{tab:sft}
\end{table}

\begin{table}[t]
    \centering
    \caption{Acc/Acc-Norm for foundational capabilities of different LLMs before and after fine-tuning.}
    \resizebox{.75\textwidth}{!}{
    \begin{tabular}{@{}l l  c c c c c}
\toprule
LLM &    & MMLU & Hellaswag & ARC & Avg. Change \\
\midrule

                  \multirow{2}{*}{Gemma-2-9b} & Before  & 71.88 &  80.08 & 64.76   
 & \multirow{2}{*}{- 0.87}  \\
                  & After  & 70.92 & 80.39  &  62.80 & \\
       \midrule
       \multirow{2}{*}{Mistral-7B-v0.3} & Before & 59.74 & 82.90 & 58.62 & \multirow{2}{*}{- 1.35}  \\
       & After & 57.56 & 82.14 &  57.51 & \\
       \midrule
      \multirow{2}{*}{Llama-3.1-8B}& Before & 67.94 & 79.33 & 55.03 & \multirow{2}{*}{+ 0.51}  \\
 & After & 66.24 & 79.48 & 58.11 & \\
\bottomrule
    \end{tabular}
    }
    \label{tab:overall_evaluation_sft}
\end{table}

\begin{table}[t]
    \centering
\caption{Performance of various LLMs on HumanEval, HumanEval+, MBPP, and MBPP+ benchmarks before and after fine-tuning.}
    \resizebox{.9\textwidth}{!}{
    \begin{tabular}{@{}l l  c c c c c c}
\toprule
LLM &    & HumanEval & HumanEval+ & MBPP & MBPP+ & Avg. Change \\
\midrule

                  \multirow{2}{*}{Gemma-2-9b} & Before  & 67.7               & 59.1                & 73.3            & 63.0
 & \multirow{2}{*}{- 0.78}  \\
                  & After  & 66.5               & 59.1                & 72.8            & 62.2            &       \\
       \midrule
       \multirow{2}{*}{Mistral-7B-v0.3} & Before & 36.0               & 31.1                & 50.3            & 42.1 & \multirow{2}{*}{- 0.68}  \\
       & After & 34.8               & 30.5                & 49.2            & 42.3  & \\
       \midrule
      \multirow{2}{*}{Llama-3.1-8B}& Before & 64.5               & 55.5                & 68.0            & 55.6   & \multirow{2}{*}{- 1.05}  \\
 & After & 63.4               & 54.3                & 67.5            & 54.2   & \\
\bottomrule
    \end{tabular}
    }
    \label{tab:code_evaluation_sft}
\end{table}

\begin{table}[t]
    \centering
    \caption{Acc/Acc-Norm for foundational capabilities of different LLMs before and after fine-tuning, where LLMs are finetuned without general-purpose datasets.}
    \resizebox{.75\textwidth}{!}{
    \begin{tabular}{@{}l l  c c c c c}
\toprule
LLM &    & MMLU & Hellaswag & ARC & Avg. Change \\
\midrule

                  \multirow{2}{*}{Gemma-2-9b} & Before  & 71.88 &  80.08 & 64.76   
 & \multirow{2}{*}{- 1.49}  \\
                  & After  & 71.41 & 79.76 & 61.09 & \\
       \midrule
       \multirow{2}{*}{Mistral-7B-v0.3} & Before & 59.74 & 82.90 & 58.62 & \multirow{2}{*}{- 1.60}  \\
       & After & 57.72 & 82.52 & 56.23 & \\
       \midrule
      \multirow{2}{*}{Llama-3.1-8B}& Before & 67.94 & 79.33 & 55.03 & \multirow{2}{*}{- 0.46}  \\
 & After & 65.65 & 80.60 & 54.68	 & \\
\bottomrule
    \end{tabular}
    }
    \label{tab:overall_evaluation_sft_exclude_general}
\end{table}

In this section, we further explore the efficacy of fine-tuning in improving the performance of LLMs.
We use the training sets described in Section \ref{exp_setup} to fine-tune three LLMs: Llama-3.1-8B \citep{llama-3.1-8b}, Gemma-2-9b \citep{gemma2}, and Mistral-7B-v0.3 \citep{mistral-7b}, with the help of the LlamaFactory framework \citep{zheng2024llamafactory} and LoRA \citep{hu2022lora}.
As shown in Table \ref{tab:eval_all}, PoT performs the best across our three datasets compared to raw instructions and CoT.
Therefore, we structure our questions to guide the LLMs in generating PoT solutions, using Python code from our datasets as expected output.
Additionally, we include general-purpose datasets (Alpaca-GPT-4 \citep{peng2023instruction} and Dolly-15k \citep{DatabricksBlog2023DollyV2}) and programming datasets (Code Alpaca \citep{codealpaca} and OpenCoder \citep{huang2024opencoder}) to improve the robustness of fine-tuned models. The inclusion of general-purpose datasets in fine-tuning is inspired by state-of-the-art coding LLMs \citep{roziere2023code, zhu2024deepseek}, which use general-purpose datasets during their training phases. The inclusion of such datasets is a common practice to ensure the models have a certain level of foundational capabilities. Hence, excluding general datasets would compromise the foundational capabilities of the models, which is shown in Table \ref{tab:overall_evaluation_sft_exclude_general}.
We continue to use the Acc as the primary evaluation metric for our datasets.
Additionally, to evaluate the foundational capabilities of the LLMs, we utilize three datasets: MMLU \citep{hendryckstest2021}, Hellaswag \citep{zellers2019hellaswag}, and ARC (AI2 Reasoning Challenge) \citep{allenai:arc}. We evaluate LLMs on these benchmarks using the LM-Evaluation-Harness framework \citep{eval-harness}, and all evaluations are conducted in a zero-shot setting. We use \textbf{Acc} or \textbf{Acc-Norm} stated in the framework as an evaluation metric.
Specifically, for Hellaswag and ARC, we use \textbf{Acc-Norm}; for MMLU, we use \textbf{Acc}.

Table \ref{tab:sft} shows the string processing capability of LLMs when prompted with PoT, while Table \ref{tab:overall_evaluation_sft} shows the foundational capabilities of LLMs and Table~\ref{tab:code_evaluation_sft} shows their performance on code generation benchmarks, both before and after fine-tuning. From these results, we can draw the following conclusions: \textbf{1) Our fine-tuning is effective.}
The results in Table \ref{tab:sft} clearly show that the performance of all three LLMs significantly improves after fine-tuning. For instance, the Mistral-7B-v0.3 \citep{mistral-7b} model's Acc on the Random String dataset improves by \textbf{60.11\%}. This finding demonstrates that our fine-tuning effectively enhances the LLMs' capability in string processing.
\textbf{2) The fine-tuned LLMs maintain their foundational capabilities.}
Table \ref{tab:overall_evaluation_sft} demonstrates that our fine-tuning has minimal impact on the foundational capabilities of LLMs. The results show that all evaluated LLMs exhibit an average performance degradation of less than \textbf{1.35\%}. In some cases, performance even improves, likely due to the inclusion of two general-purpose datasets during fine-tuning. This finding highlights that our fine-tuning not only enhances LLMs' string processing ability but also preserves their foundational capabilities. \textbf{3) The fine-tuned LLMs maintain their coding capabilities.} Results in Table \ref{tab:code_evaluation_sft} indicate that the code generation performance of LLMs shows a slight decline (no more than \textbf{1.05\%} on average) before and after fine-tuning. However, our fine-tuning method is quite naive, since it is not our primary focus. AI companies with advanced model training pipelines and substantial computational resources can better integrate our dataset with general and code-specific data, enabling them to maintain both the general and coding capabilities of LLMs.

Comparing finetuning with alternative approaches operating at the tokenizer level is another interesting direction. As a brief investigation, Table \ref{tab:eval_single_token} in Appendix shows LLMs' performance where the characters in the string are encoded separately. In this setup, each character is encoded by its token ID in the vocabulary, rather than being processed as part of a complete word or subword. Surprisingly, the performance of LLMs degrades compared to Table \ref{tab:eval_all} in the paper. One possible explanation is that this essentially employs a different tokenizer, as it encodes the same input text in a different way. The performance drops since LLMs are trained on the original tokenizer.  Retraining LLMs with new finer-grained tokenizers would be another alternative approach. Exploring such approaches would be an exciting avenue for future LLM developers, but it goes beyond the scope of our work. Additionally, retraining an LLM with a new tokenizer would require significant computational resources, which are unavailable to us. 

\section{Conclusion}
In this paper, we presented the \emph{first} comprehensive study on LLMs' capability in string processing. we proposed \emph{StringLLM}, a method to construct datasets for benchmarking string processing capability of LLMs. Following this, we used StringLLM to develop a series of datasets covering a broad range of string processing tasks and different types of strings. Extensive experiments conducted on these datasets indicated that current LLMs have limited capability in processing strings compared to humans. To address this limitation, we provided a detailed analysis to uncover the underlying reasons for LLMs' struggle with string processing tasks. Building on these insights, we proposed an effective solution that enhanced LLMs' performance via fine-tuning. Our fine-tuned models increased average test accuracy of three datasets by at least 38.80\%. Interesting future work includes 1) training more capable LLMs for string processing; 2) refining our datasets; and 3) further analyzing the underlying limitations of LLMs in string processing.

\section{Acknowledgments}
We thank the anonymous reviewers for their very constructive comments. This work was supported by the Microsoft Accelerating Foundation Models Research program as well as NSF under grant No. 2112562, 1937787, and 2125977. We also acknowledge the credits provided by Microsoft Azure.





\bibliography{iclr2025_conference}

\begin{thebibliography}{33}
\providecommand{\natexlab}[1]{#1}
\providecommand{\url}[1]{\texttt{#1}}
\expandafter\ifx\csname urlstyle\endcsname\relax
  \providecommand{\doi}[1]{doi: #1}\else
  \providecommand{\doi}{doi: \begingroup \urlstyle{rm}\Url}\fi

\bibitem[Chaudhary(2023)]{codealpaca}
Sahil Chaudhary.
\newblock Code alpaca: An instruction-following llama model for code generation.
\newblock \url{https://github.com/sahil280114/codealpaca}, 2023.

\bibitem[Chen et~al.(2021)Chen, Tworek, Jun, Yuan, Pinto, Kaplan, Edwards, Burda, Joseph, Brockman, et~al.]{chen2021evaluating}
Mark Chen, Jerry Tworek, Heewoo Jun, Qiming Yuan, Henrique Ponde De~Oliveira Pinto, Jared Kaplan, Harri Edwards, Yuri Burda, Nicholas Joseph, Greg Brockman, et~al.
\newblock Evaluating large language models trained on code.
\newblock \emph{arXiv preprint arXiv:2107.03374}, 2021.

\bibitem[Chen et~al.(2022)Chen, Ma, Wang, and Cohen]{chen2022program}
Wenhu Chen, Xueguang Ma, Xinyi Wang, and William~W Cohen.
\newblock Program of thoughts prompting: Disentangling computation from reasoning for numerical reasoning tasks.
\newblock \emph{arXiv preprint arXiv:2211.12588}, 2022.

\bibitem[Clark et~al.(2018)Clark, Cowhey, Etzioni, Khot, Sabharwal, Schoenick, and Tafjord]{allenai:arc}
Peter Clark, Isaac Cowhey, Oren Etzioni, Tushar Khot, Ashish Sabharwal, Carissa Schoenick, and Oyvind Tafjord.
\newblock Think you have solved question answering? try arc, the ai2 reasoning challenge.
\newblock \emph{arXiv:1803.05457v1}, 2018.

\bibitem[Conover et~al.(2023)Conover, Hayes, Mathur, Xie, Wan, Shah, Ghodsi, Wendell, Zaharia, and Xin]{DatabricksBlog2023DollyV2}
Mike Conover, Matt Hayes, Ankit Mathur, Jianwei Xie, Jun Wan, Sam Shah, Ali Ghodsi, Patrick Wendell, Matei Zaharia, and Reynold Xin.
\newblock Free dolly: Introducing the world's first truly open instruction-tuned llm, 2023.
\newblock URL \url{https://www.databricks.com/blog/2023/04/12/dolly-first-open-commercially-viable-instruction-tuned-llm}.

\bibitem[Costa-juss{\`a} et~al.(2022)Costa-juss{\`a}, Cross, {\c{C}}elebi, Elbayad, Heafield, Heffernan, Kalbassi, Lam, Licht, Maillard, et~al.]{costa2022no}
Marta~R Costa-juss{\`a}, James Cross, Onur {\c{C}}elebi, Maha Elbayad, Kenneth Heafield, Kevin Heffernan, Elahe Kalbassi, Janice Lam, Daniel Licht, Jean Maillard, et~al.
\newblock No language left behind: Scaling human-centered machine translation.
\newblock \emph{arXiv preprint arXiv:2207.04672}, 2022.

\bibitem[DeepSeek-AI(2024)]{deepseekai2024deepseekv2strongeconomicalefficient}
DeepSeek-AI.
\newblock Deepseek-v2: A strong, economical, and efficient mixture-of-experts language model.
\newblock \emph{arXiv preprint arXiv:2405.04434}, 2024.

\bibitem[Gao et~al.(2024)Gao, Tow, Abbasi, Biderman, Black, DiPofi, Foster, Golding, Hsu, Le~Noac'h, Li, McDonell, Muennighoff, Ociepa, Phang, Reynolds, Schoelkopf, Skowron, Sutawika, Tang, Thite, Wang, Wang, and Zou]{eval-harness}
Leo Gao, Jonathan Tow, Baber Abbasi, Stella Biderman, Sid Black, Anthony DiPofi, Charles Foster, Laurence Golding, Jeffrey Hsu, Alain Le~Noac'h, Haonan Li, Kyle McDonell, Niklas Muennighoff, Chris Ociepa, Jason Phang, Laria Reynolds, Hailey Schoelkopf, Aviya Skowron, Lintang Sutawika, Eric Tang, Anish Thite, Ben Wang, Kevin Wang, and Andy Zou.
\newblock A framework for few-shot language model evaluation, 07 2024.
\newblock URL \url{https://zenodo.org/records/12608602}.

\bibitem[Goodside(2024)]{count_straw}
Riley Goodside.
\newblock Riley's blog post, 2024.
\newblock URL \url{https://x.com/goodside/status/1830470374321963103}.

\bibitem[Google(2024)]{gemma2}
Google.
\newblock Gemma 2 is now available to researchers and developers, 2024.
\newblock URL \url{https://blog.google/technology/developers/google-gemma-2/}.

\bibitem[Hendrycks et~al.(2021)Hendrycks, Burns, Basart, Zou, Mazeika, Song, and Steinhardt]{hendryckstest2021}
Dan Hendrycks, Collin Burns, Steven Basart, Andy Zou, Mantas Mazeika, Dawn Song, and Jacob Steinhardt.
\newblock Measuring massive multitask language understanding.
\newblock \emph{Proceedings of the International Conference on Learning Representations (ICLR)}, 2021.

\bibitem[Hu et~al.(2022)Hu, Shen, Wallis, Allen-Zhu, Li, Wang, Wang, and Chen]{hu2022lora}
Edward~J Hu, Yelong Shen, Phillip Wallis, Zeyuan Allen-Zhu, Yuanzhi Li, Shean Wang, Lu~Wang, and Weizhu Chen.
\newblock Lo{RA}: Low-rank adaptation of large language models.
\newblock In \emph{International Conference on Learning Representations}, 2022.
\newblock URL \url{https://openreview.net/forum?id=nZeVKeeFYf9}.

\bibitem[Huang et~al.(2024)Huang, Cheng, Liu, Hao, Song, Xu, Yang, Liu, Zhang, Chai, et~al.]{huang2024opencoder}
Siming Huang, Tianhao Cheng, Jason~Klein Liu, Jiaran Hao, Liuyihan Song, Yang Xu, J~Yang, JH~Liu, Chenchen Zhang, Linzheng Chai, et~al.
\newblock Opencoder: The open cookbook for top-tier code large language models.
\newblock \emph{arXiv preprint arXiv:2411.04905}, 2024.

\bibitem[Meta(2024)]{llama-3.1-8b}
Meta.
\newblock Introducing llama 3.1: Our most capable models to date, 2024.
\newblock URL \url{https://ai.meta.com/blog/meta-llama-3-1/}.

\bibitem[Mistral-AI(2024{\natexlab{a}})]{codestral}
Mistral-AI.
\newblock Codestral: Hello, world!, 2024{\natexlab{a}}.
\newblock URL \url{https://mistral.ai/news/codestral/}.

\bibitem[Mistral-AI(2024{\natexlab{b}})]{mathstral-7b}
Mistral-AI.
\newblock Mathstral, 2024{\natexlab{b}}.
\newblock URL \url{https://mistral.ai/news/mathstral/}.

\bibitem[Mistral-AI(2024{\natexlab{c}})]{mistral-7b}
Mistral-AI.
\newblock Mistral 7b, 2024{\natexlab{c}}.
\newblock URL \url{https://mistral.ai/news/announcing-mistral-7b/}.

\bibitem[OpenAI(2022)]{gpt3.5}
OpenAI.
\newblock Introducing {ChatGPT}, 2022.
\newblock URL \url{https://openai.com/index/chatgpt/}.

\bibitem[OpenAI(2023{\natexlab{a}})]{achiam2023gpt}
OpenAI.
\newblock {GPT}-4 technical report.
\newblock \emph{arXiv preprint arXiv:2303.08774}, 2023{\natexlab{a}}.

\bibitem[OpenAI(2023{\natexlab{b}})]{gpt4turbo}
OpenAI.
\newblock New models and developer products announced at devday, 2023{\natexlab{b}}.
\newblock URL \url{https://openai.com/index/new-models-and-developer-products-announced-at-devday/}.

\bibitem[OpenAI(2024{\natexlab{a}})]{gpt4o}
OpenAI.
\newblock Hello {GPT-4o}, 2024{\natexlab{a}}.
\newblock URL \url{https://openai.com/index/hello-gpt-4o/}.

\bibitem[OpenAI(2024{\natexlab{b}})]{o1}
OpenAI.
\newblock Learning to reason with llms, 2024{\natexlab{b}}.
\newblock URL \url{https://openai.com/index/learning-to-reason-with-llms}.

\bibitem[Ouyang et~al.(2022)Ouyang, Wu, Jiang, Almeida, Wainwright, Mishkin, Zhang, Agarwal, Slama, Ray, et~al.]{ouyang2022training}
Long Ouyang, Jeffrey Wu, Xu~Jiang, Diogo Almeida, Carroll Wainwright, Pamela Mishkin, Chong Zhang, Sandhini Agarwal, Katarina Slama, Alex Ray, et~al.
\newblock Training language models to follow instructions with human feedback.
\newblock \emph{Advances in neural information processing systems}, 35:\penalty0 27730--27744, 2022.

\bibitem[Peng et~al.(2023)Peng, Li, He, Galley, and Gao]{peng2023instruction}
Baolin Peng, Chunyuan Li, Pengcheng He, Michel Galley, and Jianfeng Gao.
\newblock Instruction tuning with gpt-4.
\newblock \emph{arXiv preprint arXiv:2304.03277}, 2023.

\bibitem[Roziere et~al.(2023)Roziere, Gehring, Gloeckle, Sootla, Gat, Tan, Adi, Liu, Remez, Rapin, et~al.]{roziere2023code}
Baptiste Roziere, Jonas Gehring, Fabian Gloeckle, Sten Sootla, Itai Gat, Xiaoqing~Ellen Tan, Yossi Adi, Jingyu Liu, Tal Remez, J{\'e}r{\'e}my Rapin, et~al.
\newblock Code llama: Open foundation models for code.
\newblock \emph{arXiv preprint arXiv:2308.12950}, 2023.

\bibitem[Shin \& Kaneko(2024)Shin and Kaneko]{shin2024large}
Andrew Shin and Kunitake Kaneko.
\newblock Large language models lack understanding of character composition of words.
\newblock \emph{arXiv preprint arXiv:2405.11357}, 2024.

\bibitem[Tan et~al.(2024)Tan, Wei, Wang, Xie, and Huang]{tan2024can}
Zhiquan Tan, Lai Wei, Jindong Wang, Xing Xie, and Weiran Huang.
\newblock Can i understand what i create? self-knowledge evaluation of large language models.
\newblock \emph{arXiv preprint arXiv:2406.06140}, 2024.

\bibitem[Wei et~al.(2022)Wei, Wang, Schuurmans, Bosma, Xia, Chi, Le, Zhou, et~al.]{wei2022chain}
Jason Wei, Xuezhi Wang, Dale Schuurmans, Maarten Bosma, Fei Xia, Ed~Chi, Quoc~V Le, Denny Zhou, et~al.
\newblock Chain-of-thought prompting elicits reasoning in large language models.
\newblock \emph{Advances in neural information processing systems}, 35:\penalty0 24824--24837, 2022.

\bibitem[Yehudai et~al.(2024)Yehudai, Kaplan, Ghandeharioun, Geva, and Globerson]{yehudai2024can}
Gilad Yehudai, Haim Kaplan, Asma Ghandeharioun, Mor Geva, and Amir Globerson.
\newblock When can transformers count to n?
\newblock \emph{arXiv preprint arXiv:2407.15160}, 2024.

\bibitem[Zellers et~al.(2019)Zellers, Holtzman, Bisk, Farhadi, and Choi]{zellers2019hellaswag}
Rowan Zellers, Ari Holtzman, Yonatan Bisk, Ali Farhadi, and Yejin Choi.
\newblock Hellaswag: Can a machine really finish your sentence?
\newblock \emph{arXiv preprint arXiv:1905.07830}, 2019.

\bibitem[Zheng et~al.(2024)Zheng, Zhang, Zhang, Ye, Luo, Feng, and Ma]{zheng2024llamafactory}
Yaowei Zheng, Richong Zhang, Junhao Zhang, Yanhan Ye, Zheyan Luo, Zhangchi Feng, and Yongqiang Ma.
\newblock Llamafactory: Unified efficient fine-tuning of 100+ language models.
\newblock In \emph{Proceedings of the 62nd Annual Meeting of the Association for Computational Linguistics (Volume 3: System Demonstrations)}, Bangkok, Thailand, 2024. Association for Computational Linguistics.

\bibitem[Zhou et~al.(2023)Zhou, Lu, Mishra, Brahma, Basu, Luan, Zhou, and Hou]{zhou2023instruction}
Jeffrey Zhou, Tianjian Lu, Swaroop Mishra, Siddhartha Brahma, Sujoy Basu, Yi~Luan, Denny Zhou, and Le~Hou.
\newblock Instruction-following evaluation for large language models.
\newblock \emph{arXiv preprint arXiv:2311.07911}, 2023.

\bibitem[Zhu et~al.(2024)Zhu, Guo, Shao, Yang, Wang, Xu, Wu, Li, Gao, Ma, et~al.]{zhu2024deepseek}
Qihao Zhu, Daya Guo, Zhihong Shao, Dejian Yang, Peiyi Wang, Runxin Xu, Y~Wu, Yukun Li, Huazuo Gao, Shirong Ma, et~al.
\newblock {DeepSeek-Coder-V2}: Breaking the barrier of closed-source models in code intelligence.
\newblock \emph{arXiv preprint arXiv:2406.11931}, 2024.

\end{thebibliography}
\bibliographystyle{iclr2025_conference}

\appendix
\newpage
\section{Appendix}



\begin{figure}[ht]
    \centering

    \subfigure[CoT]{\includegraphics[width=\textwidth]{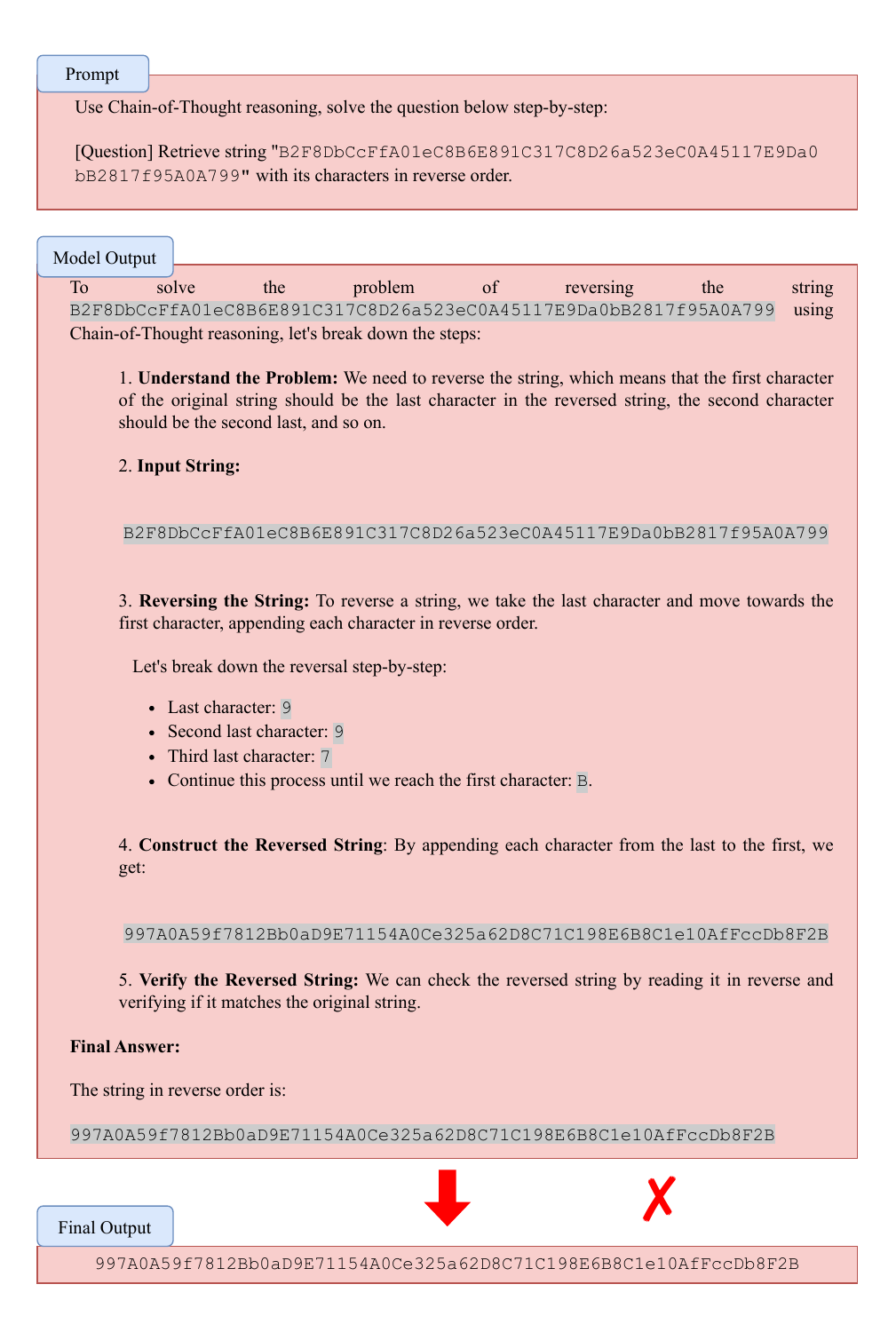}\label{fig: case_study_cot}}
    \caption{Examples of GPT-4o solving a string processing task using different prompt engineering techniques.}
    \label{fig: case_study_1}

\end{figure}

\begin{figure}[ht]
    \subfigure[Raw Instruction]{\includegraphics[width=\textwidth]{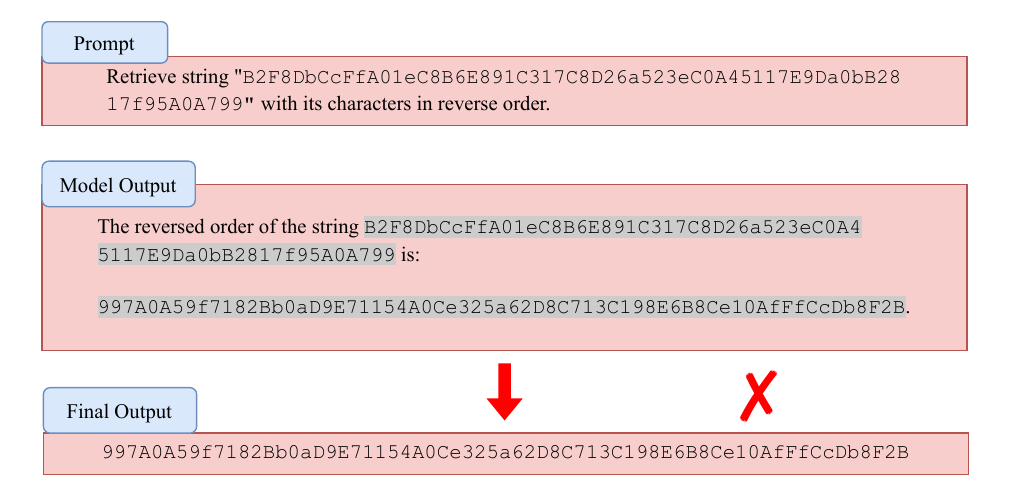}\label{fig: case_study_raw}}
    
    \subfigure[PoT]{\includegraphics[width= \textwidth]{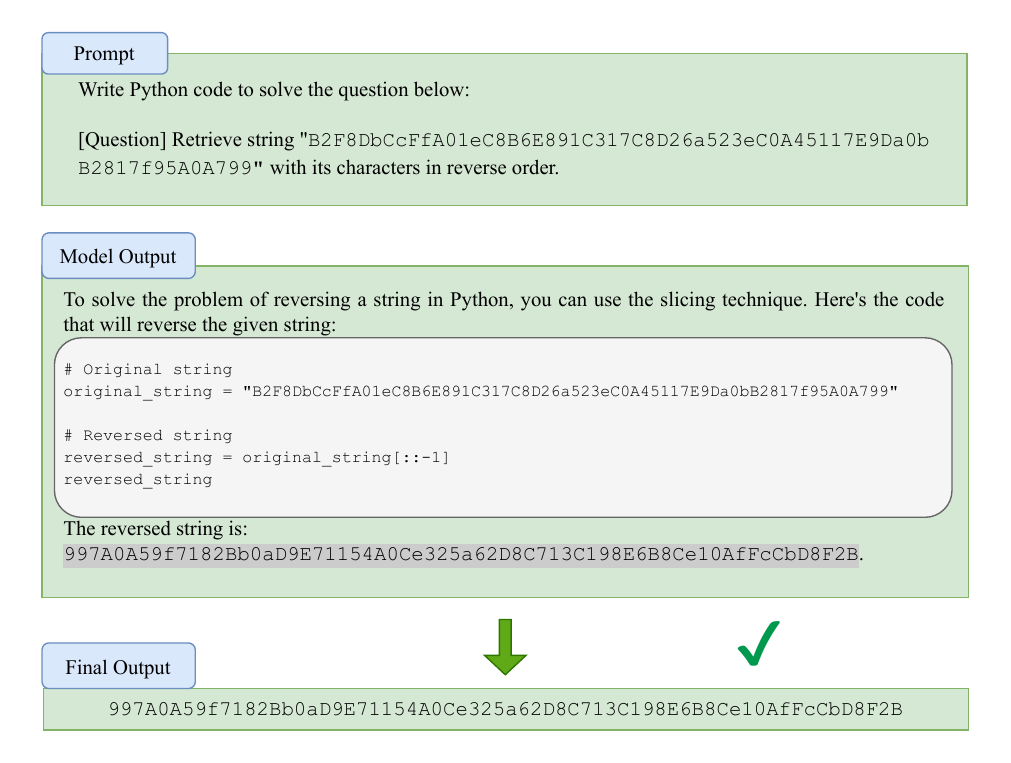}\label{fig: case_study_CoT}}
    \caption{Examples of GPT-4o solving a string processing task using different prompt engineering techniques.}
    \label{fig: case_study_2}
\end{figure}

\begin{table}[ht]
    \centering
    \renewcommand{\arraystretch}{1.2}
    \begin{tabular}{|l|c|c|}
    \hline
       LLMs  & \#Parameters  & Model Provider\\
    \hline
    \noalign{\vskip 0.8mm}
    \hline
      GPT-4-Turbo   & Unknown & OpenAI\\
      \hline
      GPT-4o   & Unknown & OpenAI\\
      \hline
      GPT-3.5   & Unknown & OpenAI\\
      \hline
      DeepSeek-Coder & 16B & DeepSeek \\
      \hline
      DeepSeek-Chat & 16B & DeepSeek\\
      \hline
      Gemma-2-9b-it & 9B & Google\\
      \hline
      Mistral-7B-v0.3 & 7B & Mistral AI\\
      \hline
      Mathstral-7B-v0.1 & 7B & Mistral AI\\
      \hline
      Codestral-22B-v0.1 & 22B & Mistral AI\\
      \hline
      Llama-3.1-8B-Instruct & 8B & Meta \\
      \hline
    \end{tabular}
    \caption{Number of parameters and model providers of LLMs used in our experiments.}
    \label{tab:2}
\end{table}

\begin{table}[ht]
\centering
\renewcommand{\arraystretch}{1.2}
\begin{tabular}{|c|c|}
\hline
Hash Function & Length \\
\hline
\noalign{\vskip 0.8mm}
\hline
MD5       & 32  \\\hline
SHA-1     & 40  \\\hline
SHA-256   & 64  \\\hline
BLAKE2b   & 128 \\\hline
SHA3-224  & 56  \\\hline
SHAKE-128  & 32  \\\hline
BLAKE2s   & 64  \\\hline
SHA3-512  & 128 \\\hline
SHAKE-256 & 64  \\\hline
SHA-384   & 96  \\
\hline
\end{tabular}
\caption{Hash functions used in our data construction and their output lengths (number of alphanumeric characters).}
\label{table:hash_lengths}
\end{table}

\begin{table}[ht]
\centering
\renewcommand{\arraystretch}{1.2}
\resizebox{\textwidth}{!}{
\begin{tabular}{@{}l|c|l@{}}
\hline
Code & Variables & Templates\\
\hline
\noalign{\vskip 0.8mm}
\hline
\func{ans = a + b}    & \func{a,b}  & Concat string \{a\} and \{b\}. \\\hline
\func{ans = a * b}       & \func{a,b}  &  Repeat string \{a\} for \{b\} times.\\\hline
\func{ans = a[:y]}  & \func{a,y} & Retrieve the first \{y\} characters from string \{a\}. \\\hline
\func{ans = a[::-1]} & \func{a} & Get the reverse of the string \{a\}. \\\hline
\func{ans = len(a)}    & \func{a}  & Determine the number of characters in the string \{a\}.\\\hline
\func{ans = x in y} & \func{x,y}  & Does string \{y\} contain substring \{x\}? \\\hline
\func{ans = a.count(x)}   & \func{a,x} & Check the occurrence of \{x\} within string \{a\}. \\\hline
\func{ans = a.strip(x)}   & \func{a,x} & Remove characters in \{x\} from both ends of string \{a\}.\\\hline
\func{ans = a.startswith(x)}  & \func{a,x} & Identify if string \{a\} starts with \{x\}. \\\hline
\func{ans = a.endswith(x)}   & \func{a,x} & Determine if string \{a\} ends with substring \{x\}. \\
\hline
\end{tabular}
}
\caption{Examples of atomic tasks.}
\label{table:atomic_tasks}
\end{table},

\begin{table}[t]
    \centering
    \caption{Accuracy for string processing capability of different LLMs, where each character of strings is assigned a single token ID.}
    \label{tab:eval_single_token}
    \resizebox{.75\textwidth}{!}{
    \begin{tabular}{@{}l c  c c c c c}
\toprule
LLM & Method  & Multilingual & Hash & Random String & AVG \\
\midrule
\multirow{3}{*}{Gemma-2-9b}& Raw Inst. & 18.46 & 19.69 & 12.34	& 16.83
 \\
                  & CoT & 20.55 & 23.59 & 16.23 & 20.12 

\\
                  & PoT & 50.86 & 54.35 & 16.95 & 40.72  \\

       \midrule
\multirow{3}{*}{Mistral-7B-v0.3}& Raw Inst. & 4.37 & 4.68 & 2.82 & 3.96
 \\
                  & CoT & 10.19 & 10.98 & 5.22 & 8.80 

\\
                  & PoT & 29.18 & 30.63 & 13.31 & 24.37  \\
       \midrule

\multirow{3}{*}{Llama-3.1-8B}& Raw Inst. & 9.62 & 17.09 & 8.11 & 11.61  \\
                  & CoT & 16.76 & 22.32 & 18.14 & 19.07 \\
                  & PoT &  38.20 & 39.06 & 17.74 & 31.67
 \\
\bottomrule

    \end{tabular}
    }
    
\end{table}

\end{document}